\documentclass[letterpaper, 10 pt, conference]{ieeeconf}  % Comment this line out if you need a4paper

\usepackage{url}

\usepackage{graphicx}

\usepackage{cite}

\usepackage{multirow}

\usepackage{amsmath}

\usepackage{array}

\usepackage{graphicx}

\usepackage{adjustbox}

\usepackage{amsmath,amssymb}

\usepackage{balance}

\usepackage{algorithm}
\usepackage{algorithmic}
\usepackage{booktabs}
\usepackage{multirow}

\IEEEoverridecommandlockouts                              % This command is only needed if 
\title{\LARGE \bf
Exploiting Spatiotemporal Properties for Efficient Event-Driven Human Pose Estimation}

\author{
Haoxian Zhou$^{1*}$, Chuanzhi Xu$^{1*}$, Langyi Chen$^{1}$, Pengfei Ye$^{2}$, Haodong Chen$^{1}$, Yuk Ying Chung$^{1}$, Qiang Qu$^{1}$
\thanks{$^{1}$ The authors are with The University of Sydney, NSW, Australia
(* Corresponding authors with e-mails: hzho0442@uni.sydney.edu.au, chuanzhi.xu@sydney.edu.au).}
\thanks{$^{2}$ Pengfei Ye is with the Hong Kong University of Science and Technology, Hong Kong, China.}
}

\begin{document}

\maketitle
\thispagestyle{empty}
\pagestyle{empty}

%%%%%%%%%%%%%%%%%%%%%%%%%%%%%%%%%%%%%%%%%%%%%%%%%%%%%%%%%%%%%%%%%%%%%%%%%%%%%%%%
\begin{abstract}

Human pose estimation focuses on predicting body keypoints to analyze human motion. Currently, most pose estimation tasks rely on conventional RGB cameras. In contrast, event cameras provide high temporal resolution and low latency, enabling robust estimation under challenging conditions and opening up new possibilities for pose estimation. However, most existing methods convert event streams into dense event frames, which adds extra computation and sacrifices the high temporal resolution of the event signal.
In this work, we aim to exploit the spatiotemporal properties of event streams based on point cloud-based framework, designed to enhance human pose estimation performance while maintaining computational efficiency. We design Event Temporal Slicing Convolution module to capture short-term dependencies across event slices, and combine it with Event Slice Sequencing module for structured temporal modeling. We further propose an edge-enhanced point cloud-based event representation to enhance spatial edge information under sparse event conditions to further improve performance.
Experiments on the DHP19 dataset show our proposed method consistently improves performance across three representative point cloud backbones: PointNet, DGCNN, and Point Transformer, with an average MPJPE reduction of 4\%.

\end{abstract}

%%%%%%%%%%%%%%%%%%%%%%%%%%%%%%%%%%%%%%%%%%%%%%%%%%%%%%%%%%%%%%%%%%%%%%%%%%%%%%%%

\section{INTRODUCTION}

Human Pose Estimation (HPE) aims to predict keypoint positions from visual inputs to analyze body structure and motion. It is a critical task in robotics and computer vision, as it enables robots to understand human motion and body structures. Accurately perceiving human poses is essential for many robotic applications, such as human-robot collaboration, action recognition, human-computer interaction, etc \cite{[1]}. Although deep learning methods using standard RGB cameras have made great progress \cite{[2], [3]}, they often struggle in challenging real-world robotic scenarios. For instance, in high-speed industrial tasks or low-light environments, traditional cameras suffer from motion blur and limited dynamic range. These hardware constraints can lead to failures in a robot’s perception and decision-making during fast movements.

Event cameras are bio-inspired asynchronous sensors that capture per-pixel brightness changes at microsecond resolution. Event cameras provide microsecond-level temporal resolution and minimal power consumption \cite{[4]}. These properties have enabled strong performance in action recognition \cite{[action]}, object tracking \cite{[tracking]}, 3D reconstruction \cite{[o1]} and super resolution \cite{[o2]}. Event cameras can overcome the limitations of frame-based cameras in extreme scenarios, enabling stable pose estimation. However, the sparse and asynchronous nature of event streams presents a significant challenge: how to effectively represent and model this data for structural tasks without losing their inherent advantages.

\begin{figure}[!t]
\centerline{\includegraphics[width=\columnwidth]{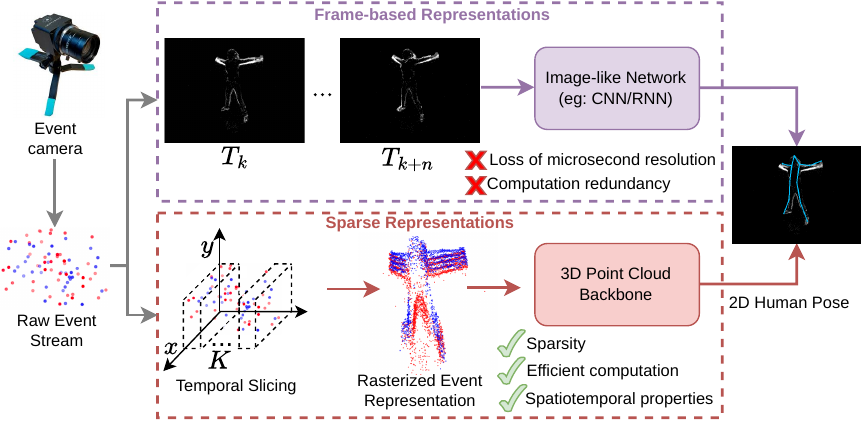}}
\caption{Comparison of two representation methods for event-based HPE. Frame-based methods aggregate events into fixed-rate dense frames, which sacrifices microsecond temporal resolution and introduces computational redundancy. The proposed method represents event streams as 3D point clouds, preserving the inherent sparsity for more efficient computation.}
\label{fig:intro}
\end{figure}

Most existing approaches convert event streams into dense frames to leverage standard vision backbones, this process often sacrifices the inherent sparsity and temporal precision of the signal. Recent attempts to represent events as point clouds have improved computational efficiency, and a comparison of these two paradigms is illustrated in Fig.~\ref{fig:intro}. 

However, they primarily focus on spatial geometry while leaving the dynamic temporal correlations between events under-exploited. We observe that human motion is inherently continuous. Although a single temporal slice may contain fragmented event data due to the sensor's sparse nature, the motion cues residing in adjacent slices are vital for maintaining pose consistency. Motivated by this, we propose a spatio-temporal framework that explicitly bridges these temporal gaps within a point cloud-based pipeline. Experimental results on the DHP19 dataset verify that our method consistently improves three representative point cloud backbones, namely PointNet, DGCNN, and Point Transformer, yielding an average MPJPE reduction of 4\%.

\begin{figure*}[ht]
    \centering
    \includegraphics[width=\textwidth]{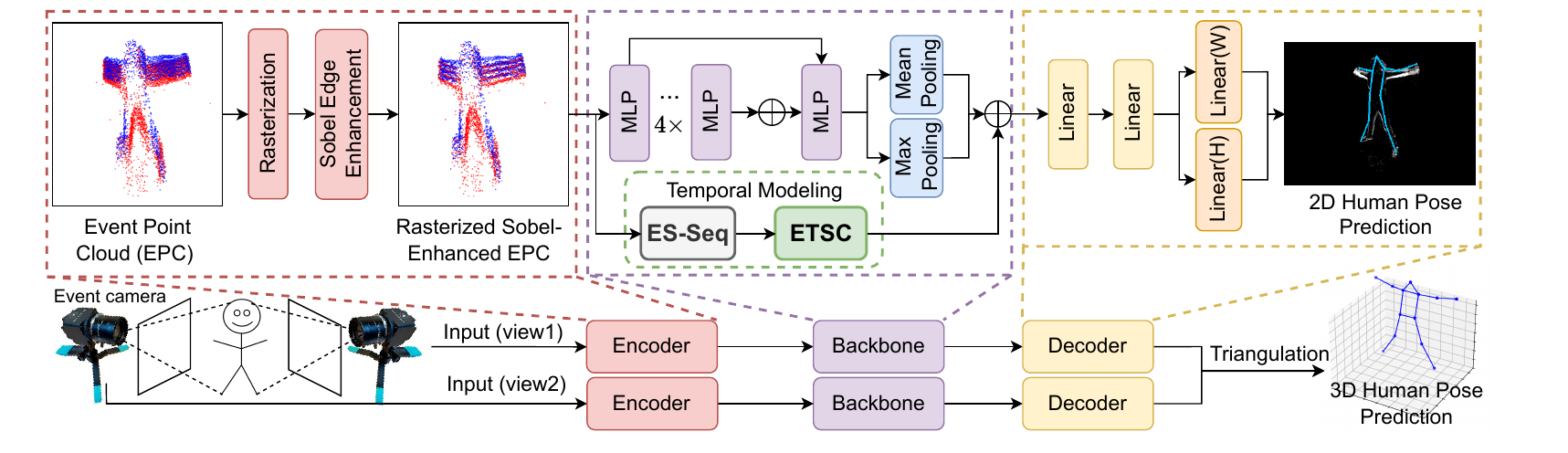} 
    \caption{Overview of the proposed pipeline. Event point clouds are rasterized with Sobel-based spatial edge enhancement and fed into the backbone. Temporal modeling with ES-Seq and ETSC refines the features, and SimDR \cite{[SimDR]} decodes 2D poses from each view, which are triangulated into the final 3D pose. }
    \label{fig:pipeline}
\end{figure*}

Our contributions can be summarized as follows:
\begin{itemize}
    \item We propose Event Temporal Slicing Convolution (ETSC) module to capture short-term temporal dependencies across event slices, and integrate it into point cloud networks to fully exploit the sparsity and temporal features of event data for human pose estimation.
    \item We design Event Slice Sequencing (ES-Seq) module that assigns event point clouds along the temporal dimension to provide temporally structured sequences.
    \item We introduce an event Sobel edge enhancement module to strengthen spatial edge features in events and improve the model’s ability to perceive motion boundaries.

\end{itemize}

\section{RELATED WORK}

\subsection{Human Pose Estimation}
Human Pose Estimation initially relied on handcrafted features. DeepPose \cite{[HPE1]} introduced convolutional neural networks (CNNs) to HPE, marking the beginning of deep learning–based approaches in this field. Subsequently, the Stacked Hourglass network \cite{[HPE6]} established the dominance of heatmap representations for capturing spatial uncertainty through repeated bottom-up and top-down inference. In multi-person pose estimation, OpenPose \cite{[HPE7]} addressed the keypoint association problem in bottom-up pipelines by introducing Part Affinity Fields (PAFs). For 3D and video settings, VideoPose3D \cite{[HPE2]} employed dilated temporal convolutions to model sequence information. With the advent of the Transformer era, TokenPose \cite{[HPE3]} adopted keypoint tokenization for structured pose modeling. More recently, DiffPose \cite{[HPE4]} utilized denoising diffusion models to handle multi-hypothesis uncertainty, while UniPose \cite{[HPE5]} incorporated large language models (LLMs) into the HPE framework, opening new directions for unified pose understanding. Despite the progress of RGB-based methods, conventional frame cameras remain limited in challenging scenarios such as fast motion or low-light environments. This motivates the use of event cameras that asynchronously capture brightness changes with microsecond temporal resolution.

\subsection{Frame-based Representations in Event HPE}
Most event-driven human pose estimation methods rely on deep learning. Researchers have explored various event representations \cite{[5], [6], [7]}, such as event frames, to handle event sparsity as input to CNNs. Calabrese et al. introduced the first event-based human pose estimation dataset \cite{[3HPE1]} with an event frame-based baseline method. Xu et al. combined event frames with grayscale images for 3D human motion capture \cite{[3HPE2]}, but required heavy optimization. Later, Zou et al. estimated optical flow directly from events \cite{[3HPE3]}, which aimed to reduce grayscale dependence in \cite{[3HPE2]}. Scarpellini et al. proposed the first monocular event camera approach for 3D pose estimation across three orthogonal planes \cite{[3HPE4]}. More recently, some methods attempted to address the problem of pose disappearance in static scenes. Goyal et al. \cite{[2HPE1]} introduced a novel event representation, Exponentially Reduced Ordinal Surface \cite{[8]}, to mitigate the temporal decay. Shao et al. \cite{[2HPE2]} used a Long Short-Term Memory \cite{[LSTM]} network to recover missing information. However, most of the above approaches convert event streams into dense frames, which destroys sparsity and adds redundant computation. Furthermore, it is difficult to preserve dynamic temporal details for human pose estimation under such fixed frame-rate constraints.

\subsection{Sparse Representations in Event HPE}
Only a few methods have exploited the sparse representation of events. 
Zou et al. converted event streams into spike-based representations \cite{[3HPE10]} and employed spiking neural networks \cite{[SNN]} to exploit event sparsity. 
However, this representation has limited capability to model the spatial geometric relationships of events.
Chen et al. represented event streams as event point clouds and processed them with point cloud networks \cite{[3HPE5]}, achieving a balance between computational efficiency and spatial representation.

Despite these advantages, existing point cloud-based approaches \cite{[3HPE5]} mainly focus on spatial geometric modeling while overlooking the temporal properties of event streams. Since events are triggered by brightness changes, body parts that remain static do not generate events, which may lead to incomplete spatial observations. This issue is particularly evident within short temporal windows, where important motion cues may be distributed across adjacent slices.

\begin{algorithm}[t]
\caption{Temporal Modeling with ES-Seq and ETSC}
\label{alg:tm}
\begin{algorithmic}[1]
\REQUIRE Point features $feat \in \mathbb{R}^{B\times C\times N}$, normalized timestamps $\tau \in [0,1]^{B\times N}$, slice number $K$
\STATE \textbf{// ES-Seq: build slice tokens from unstructured points}
\STATE $slice\_id \leftarrow \lfloor \tau \cdot K \rfloor,\ \ slice\_id \in \{0,\dots,K-1\}$
\STATE $g_{\max} \leftarrow \mathrm{GlobalMaxPool}(feat)$ \COMMENT{for empty-slice fallback}
\FOR{$s=0$ \textbf{to} $K-1$}
    \STATE $t_s \leftarrow \max\limits_{n:\,slice\_id(n)=s} feat[:,:,n]$ \COMMENT{intra-slice max pooling}
    \IF{slice $s$ is empty}
        \STATE $t_s \leftarrow g_{\max}$
    \ENDIF
\ENDFOR
\STATE $\mathbf{T} \leftarrow [t_1,t_2,\dots,t_K] \in \mathbb{R}^{B\times K\times C}$

\STATE \textbf{// ETSC: temporal convolution on slice sequence}
\STATE $\mathbf{T}' \leftarrow \mathrm{Conv1D}(\mathbf{T}; k{=}3, d{=}1)$
\STATE $\mathbf{T}' \leftarrow \mathrm{Conv1D}(\mathbf{T}'; k{=}3, d{=}2)$
\STATE $\mathbf{T}' \leftarrow \mathrm{ReLU}(\mathrm{BN}(\mathbf{T}' + \mathbf{T}))$ \COMMENT{residual}

\STATE \textbf{// temporal global descriptor}
\STATE $t_{\text{global}} \leftarrow \mathrm{MeanPool}_{K}(\mathbf{T}') \in \mathbb{R}^{B\times C}$
\RETURN $t_{\text{global}}$
\end{algorithmic}
\end{algorithm}

\section{Method}

To address the limitation that existing point cloud-based methods insufficiently exploit temporal dependencies in event streams, we propose a spatial edge-enhanced point cloud-based framework with explicit cross-slice temporal modeling. Fig.~\ref{fig:pipeline} shows the overall pipeline of our proposed method, where we use PointNet serving as a representative backbone. To achieve 3D triangulation \cite{[3HPE1]}, at least two event cameras are required to capture the same object. The two cameras are placed at an oblique angle of approximately 90° to form a frontal binocular disparity. The dual-view input is then fed into the pipeline for processing. In the following, we describe our method in detail.

\subsection{Rasterized Event Representation}
An event camera produces an asynchronous and sparse event stream $(x,y,t,p)$, where $(x,y)$ denotes the pixel coordinates, $t$ is the microsecond-level timestamp, and $p$ represents the brightness changes. We use the rasterized event point cloud representation proposed in [21] to leverage the sparse and spatiotemporal properties of events, instead of event frames. 

Event cameras generate events at microsecond-level temporal resolution, and directly feeding all events into the network is computationally expensive. To reduce events while preserving information, we accumulate them on a pixel grid. Specifically, for a given time window $[T_i,T_{i+1})$, we divide it into $K$ equal-length sub-segments. We find that $K=4$ achieves a suitable balance between maintaining temporal resolution and reducing computational complexity. For each time slice, we aggregate events on a pixel grid. If multiple events ${(t_m,p_m)}_{m=1}^M$ fall into the same pixel $(x,y)$, we compute:
\begin{equation}
t_{\text{avg}}=\frac{1}{M}\sum_{m=1}^{M}t_m,\quad
p_{\text{acc}}=\sum_{m=1}^{M}p_m,\quad
e_{\text{cnt}}=M,
\end{equation}
where $t_{\text{avg}}$ denotes the average timestamp of events within the slice (normalized to $[0,1]$), $p_{\text{acc}}$ represents the accumulated polarity, and $e_{\text{cnt}}$ denotes the event count. Finally, each valid pixel corresponds to a 5-dimensional point:
\begin{equation}
(x,\;y,\;t_{\text{avg}},\;p_{\text{acc}},\;e_{\text{cnt}}).
\end{equation}

\begin{figure}[!t]
\centerline{\includegraphics[width=\columnwidth]{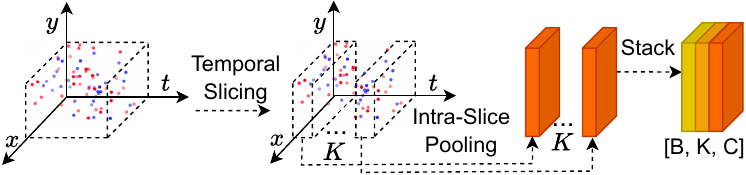}}
\caption{Structure of the Event Slice Sequencing (ES-Seq) module.}
\label{fig:esS}
\end{figure}

\subsection{Spatial Edge-Enhanced Event Representation}
Regarding spatial enhancement of events, Xu et al. \cite{[o3]} introduced a Sobel Event Frame representation to enhance spatial gradients for event-based 3D reconstruction. Following a similar motivation, we propose an event edge enhancement module that applies Sobel convolution in the voxel grid domain, making it compatible with the rasterized event point cloud representation.

Since events are triggered by brightness changes, the polarity $p_{\text{acc}}$ directly reflects the sign of these changes, acting like a gradient direction signal. We aim to enhance $p_{\text{acc}}$ to strengthen spatial edges and help the network better localize body parts under sparse event conditions.

First, we introduce an edge enhancement mechanism based on Sobel convolution in the voxel grid domain. Specifically, for each temporal slice, we construct an event count map $e_{\text{cnt}}(x,y)$ from accumulated events. We apply the classical Sobel operator to compute horizontal and vertical gradients:
\begin{equation}
G_x = \text{Conv2D}(e_{\text{cnt}}, K_x), \quad
G_y = \text{Conv2D}(e_{\text{cnt}}, K_y),
\end{equation}
where $K_x$ and $K_y$ are Sobel kernels. The edge magnitude is then obtained as:
\begin{equation}
E(x,y)=\sqrt{G_x(x,y)^2+G_y(x,y)^2}.
\end{equation}
Then, $E(x,y)$ is normalized by the maximum value within the same slice to ensure scale invariance across channels:
\begin{equation}
\tilde{E}(x,y)=\frac{E(x,y)}{\max E(x,y)+\varepsilon},
\end{equation}
where $\max E(x,y)$ indicates the maximum edge magnitude over the entire slice.

Next, we construct the enhancement weight:
\begin{equation}
w(x,y)=1+\alpha \cdot \tilde{E}(x,y),
\end{equation}
where $\alpha\in[0,1]$ controls the enhancement strength (set it at 0.5 in the experiment).

Finally, we perform per-pixel modulation on the selected statistic, the polarity accumulation $p_{\text{acc}}$:
\begin{equation}
p'_{\text{acc}}=w\cdot p_{\text{acc}}.
\end{equation}

The enhancement increases edge responses in the voxel grid stage, and the enhanced statistics are then exported as point cloud representations.

\begin{figure}[!t]
\centerline{\includegraphics[width=\columnwidth]{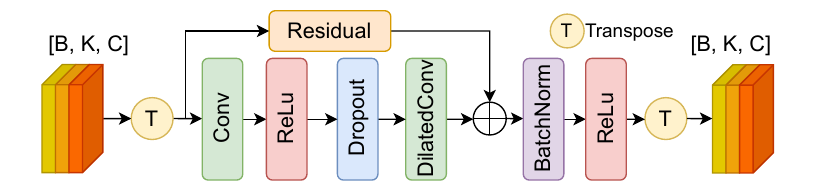}}
\caption{Structure of the Event Temporal Slice Convolution (ETSC) module.}
\label{fig:etsc}
\end{figure}

\subsection{Temporal Modeling}
To exploit the sparse structure of event point clouds, we perform temporal modeling (See Alg. \ref{alg:tm})
First, we introduce the ES-Seq module as a novel representation that organizes unstructured event points into structured short-term temporal sequences, as shown in Fig.~\ref{fig:esS}.
First, the module maps each event point cloud into slices according to the normalized timestamp $t_{\text{avg}} \in [0,1]$:

\begin{equation}
\text{slice\_id} = \lfloor t_{\text{avg}} \cdot K \rfloor, \quad \text{slice\_id}\in[0, K-1],
\end{equation}
where $K$ is the number of temporal slices and $\text{slice\_id}$ is the discrete slice index (time-bin ID) computed from $t_{\text{avg}}$. In this way, each point is assigned to a fixed slice. 

Within each slice, ES-Seq extracts point-level features from all points using max pooling to obtain a token representation:
\begin{equation}
t_{s,c} = \max_{n \in \text{slice } s}  feat_{c,n},
\end{equation}
where $s$ is the temporal slice index (1-$K$), $c$ is the channel dimension, $n$ is the point index within the slice, and $feat$ is the per-point latent feature representation of events.

Finally, the $K$ slice tokens are stacked in temporal order to form a regularized short-term sequence:
\begin{equation}
\mathbf{T}=[t_1,t_2,\dots,t_K]\in \mathbb{R}^{B\times K\times C}.
\end{equation}

Bai et al. \cite{[TCN]} demonstrated the effectiveness of dilated 1D convolution for sequence modeling. However, their design mainly targets dense and long sequential data, which limits its applicability to event streams. In contrast, our proposed ETSC module operates on slice-level tokens rather than frame-level sequences and uses a dilated convolution and residual design optimized for ultrashort event sequences, enabling efficient short-term temporal modeling under sparse conditions.
We feed the short-term sequence into ETSC to capture local motion patterns and temporal dependencies:
\begin{equation}
\mathbf{T}' = \text{ETSC}(\mathbf{T}), \quad \mathbf{T}' \in \mathbb{R}^{B \times K \times C}.
\end{equation}

The structure of ETSC is shown in Fig.~\ref{fig:etsc}. Specifically, the module extracts local dependencies across slices using a standard convolution layer and a dilated convolution layer. In our implementation, these two layers employ dilation rates of 1 and 2 respectively, both with a kernel size of 3.

After obtaining the temporal token representations, we take the mean along the temporal dimension to obtain a global temporal vector:
\begin{equation}
t_{\text{global}}=\frac{1}{K}\sum_{i=1}^K t'_i \in\mathbb{R}^{B\times C}.
\end{equation}
Then, we concatenate it with $g_{\text{max}}$ and $g_{\text{avg}}$:
\begin{equation}
g_{\text{all}} = [g_{\text{max}}; g_{\text{avg}}; t_{\text{global}}] \in \mathbb{R}^{B \times 3C},
\end{equation}
where $g_{\text{max}}$ and $g_{\text{avg}}$ are obtained from $feat$ by global max pooling and global average pooling over points.

As shown in the backbone of Fig.~\ref{fig:pipeline}, we first extract point-wise features feat from the point cloud backbone. Then, using the input timestamp $t_{\text{avg}}$, we divide the points into $K$ temporal slices by ES-Seq. Within each slice, pooling is applied to obtain a sequence of slice tokens. The ETSC module is then used to model short-term temporal dependencies, producing a global temporal representation $t_{\text{global}}$. Finally, $t_{\text{global}}$ is concatenated with the global spatial features ($g_{\text{max}}$, $g_{\text{avg}}$) and fed into the task head for keypoint regression.

\begin{figure*}[t]
\centering
\setlength{\tabcolsep}{1pt}      
\renewcommand{\arraystretch}{0.1}

\resizebox{0.85\textwidth}{!}{%
\begin{tabular}{ccccc}

\rotatebox{90}{\parbox{3cm}{\centering PointNet}} &
\includegraphics[width=0.23\textwidth,height=0.17\textwidth]{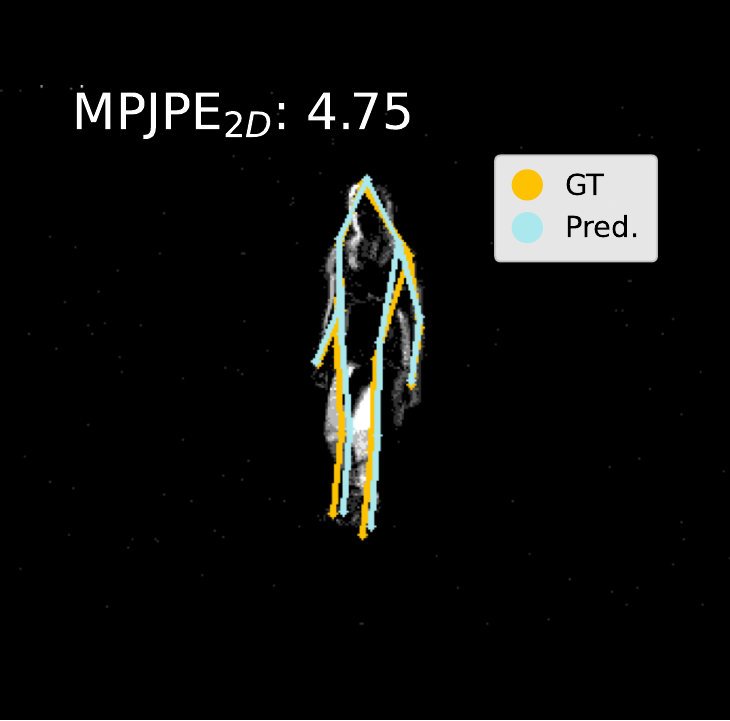}  &
\includegraphics[width=0.23\textwidth,height=0.17\textwidth]{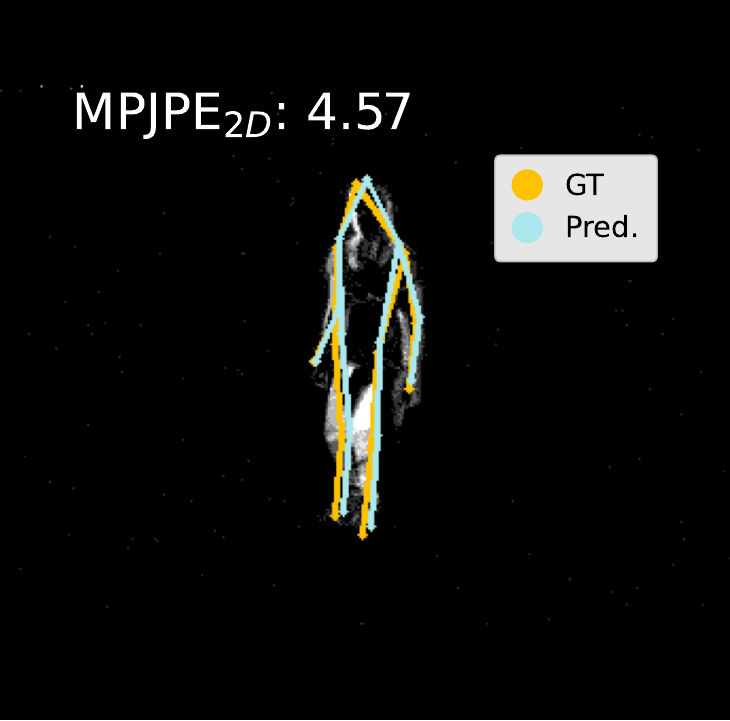} &
\includegraphics[width=0.23\textwidth,height=0.17\textwidth]{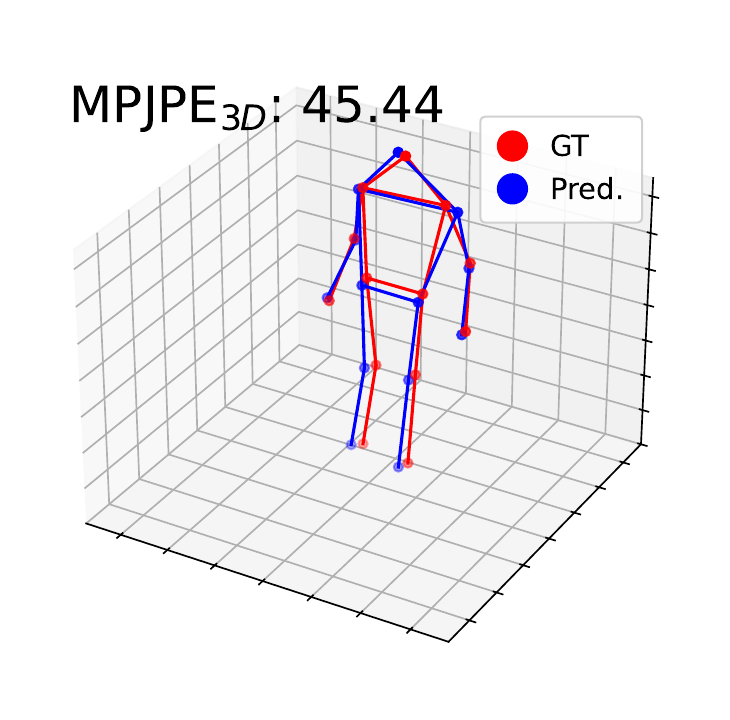}  &
\includegraphics[width=0.23\textwidth,height=0.17\textwidth]{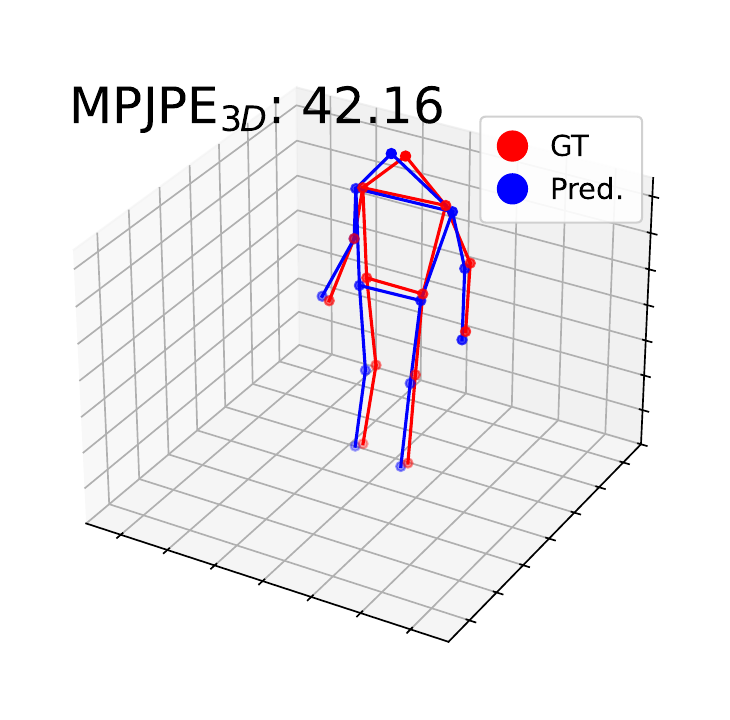} \\

\rotatebox{90}{\parbox{3cm}{\centering DGCNN}} &
\includegraphics[width=0.23\textwidth,height=0.17\textwidth]{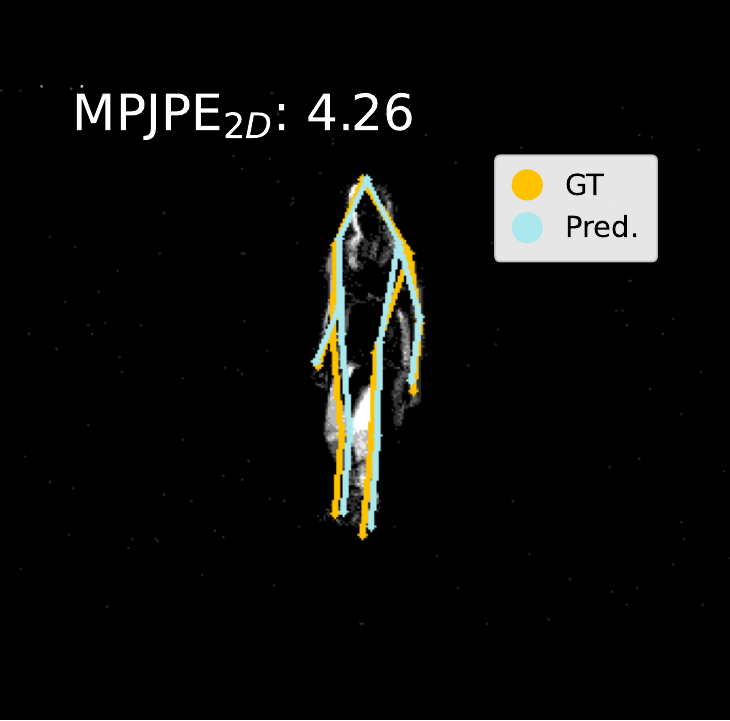}  &
\includegraphics[width=0.23\textwidth,height=0.17\textwidth]{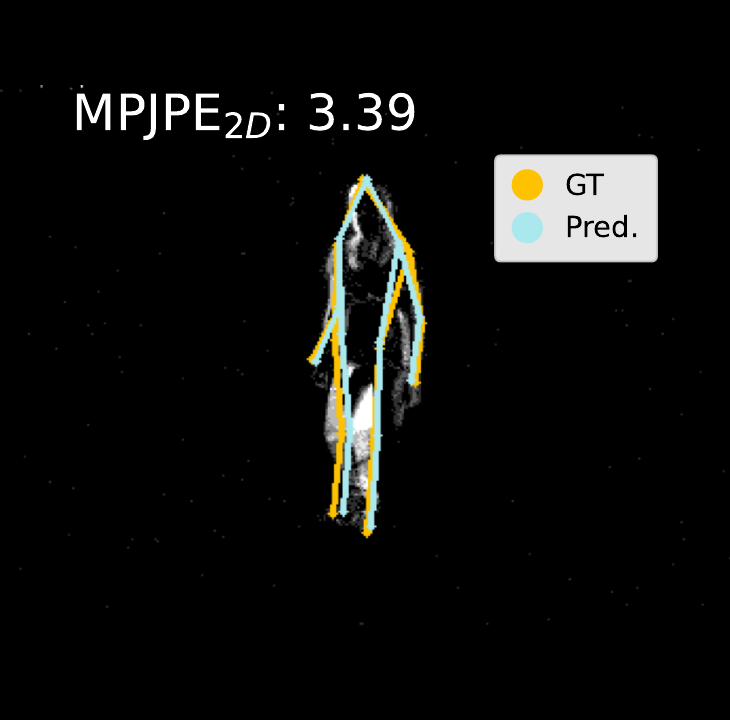} &
\includegraphics[width=0.23\textwidth,height=0.17\textwidth]{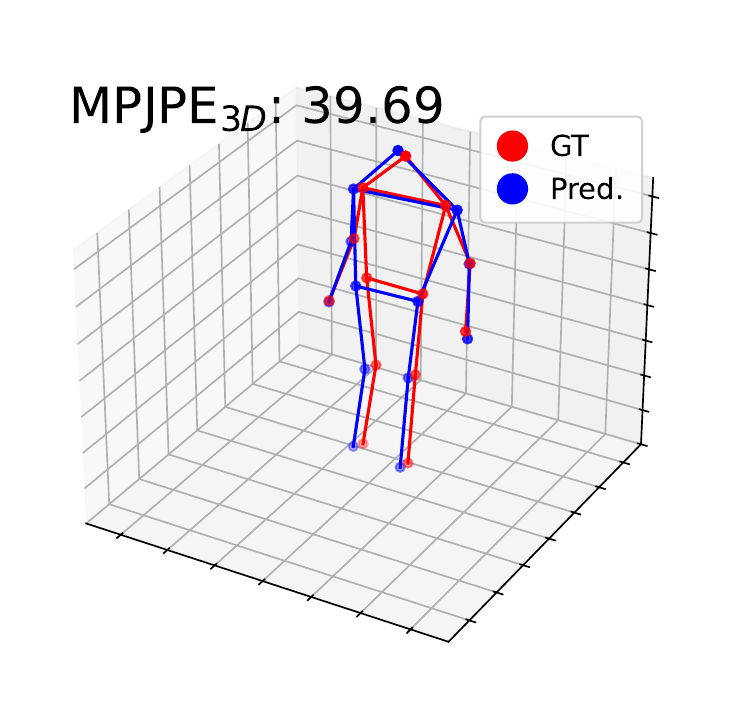}  &
\includegraphics[width=0.23\textwidth,height=0.17\textwidth]{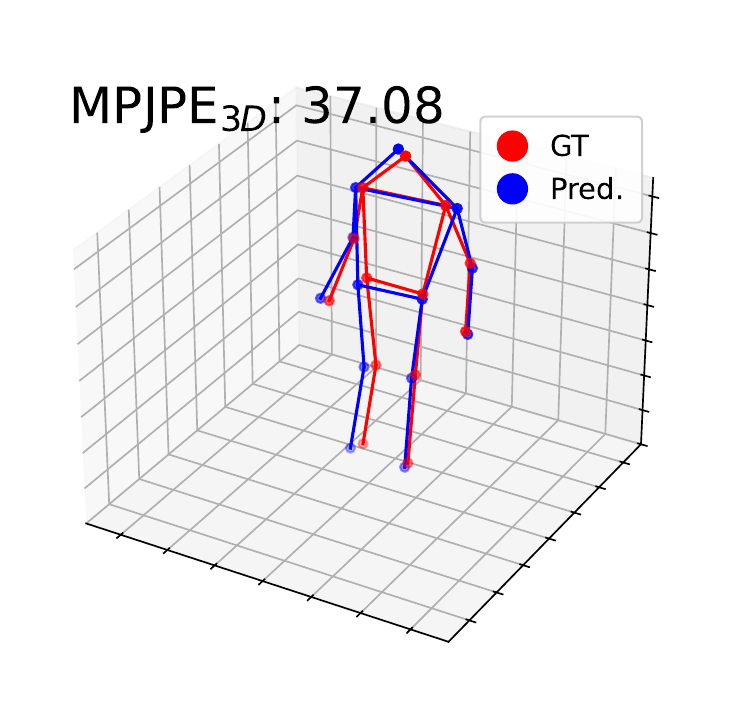} \\

\rotatebox{90}{\parbox{3cm}{\centering PointTrans}} &
\includegraphics[width=0.23\textwidth,height=0.17\textwidth]{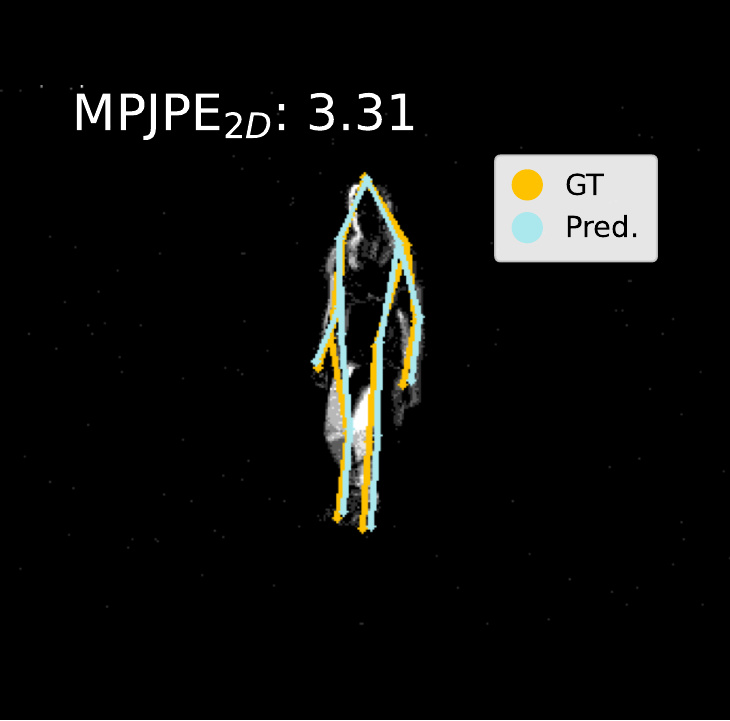}  &
\includegraphics[width=0.23\textwidth,height=0.17\textwidth]{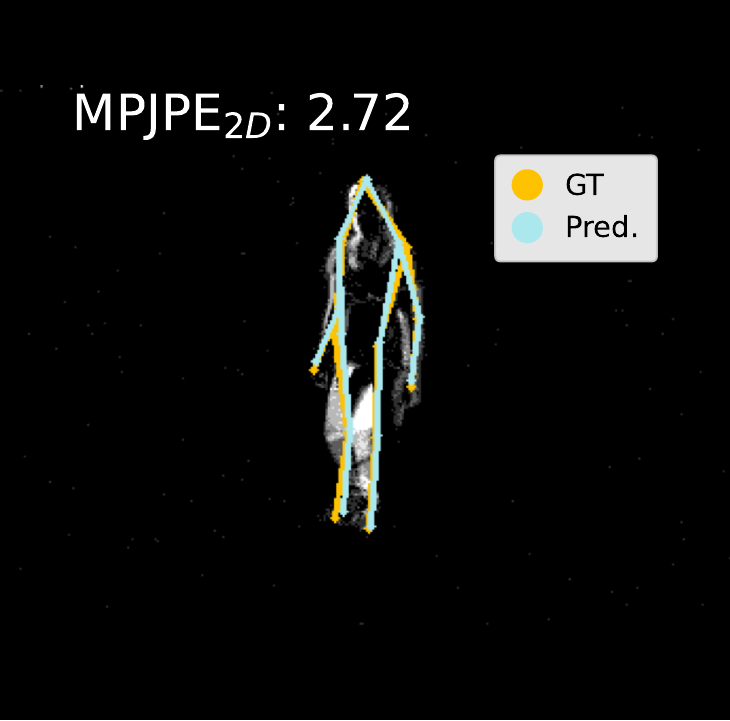} &
\includegraphics[width=0.23\textwidth,height=0.17\textwidth]{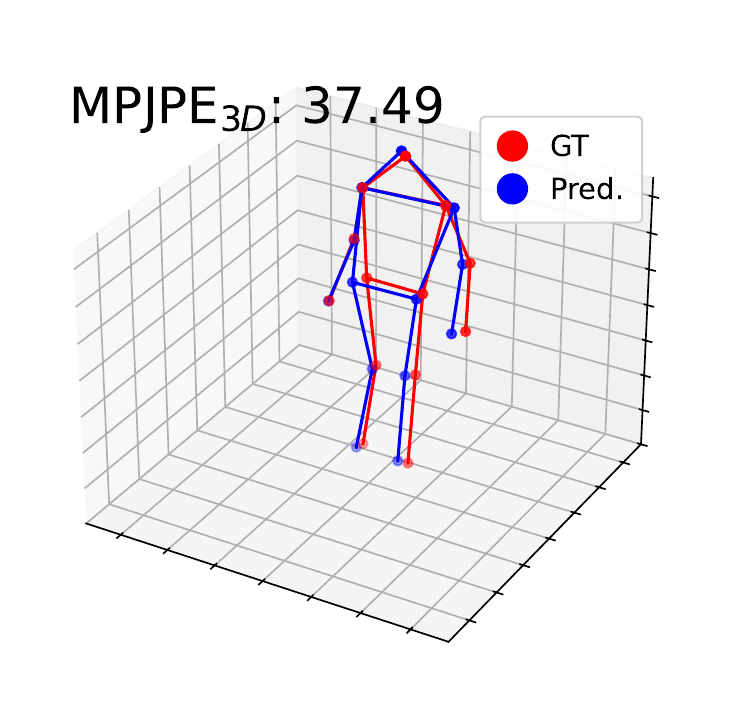}  &
\includegraphics[width=0.23\textwidth,height=0.17\textwidth]{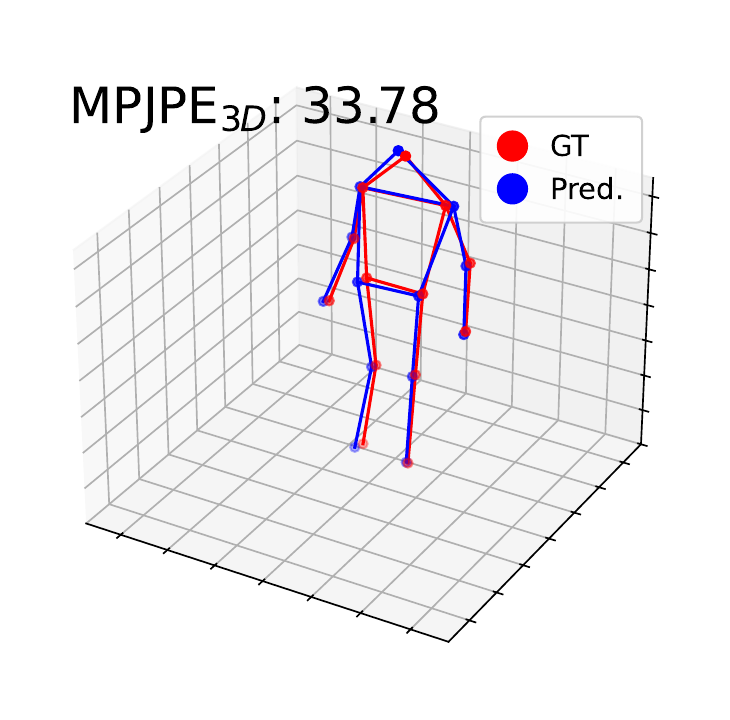} \\

\\[1mm] 
& \small (a) Baseline (2D)
& \small (b) Ours (2D)
& \small (c) Baseline (3D)
& \small (d) Ours (3D) \\
\end{tabular}
}

\caption{Visualization of results from different models on the DHP19 dataset. (a–b) 2D results from cam3 view (baseline vs. ours). (c–d) 3D results via triangulation from cam2 \& cam3 views (baseline vs. ours).}
\label{fig:result}
\end{figure*}

\section{Experiment}

\subsection{Datasets}
DHP19 \cite{[3HPE1]} is the only public event-based human pose estimation dataset that provides raw event streams, captured by four event cameras placed around the subject to provide 360° coverage, with 17 participants and 33 actions. The MMHPSD \cite{[3HPE3]} was also captured using real event cameras, but it includes limited motion types and only publicly provides frame-based data, which cannot be reconstructed into the original event streams for point cloud rasterization. Therefore, we primarily use DHP19 in this work.

In addition, we employ the Event-Human3.6M dataset \cite{[h36m]} for inference visualization to further demonstrate the generalization ability of our model. Event-Human3.6M \cite{[2HPE1]} is generated from the widely used Human3.6M dataset by converting RGB videos into event streams using the v2e simulator \cite{[v2e]}. The dataset is designed for single-person 2D HPE and provides event sequences with a spatial resolution of $640\times480$ and pose annotations at 50 Hz. It contains recordings from 11 subjects performing 17 different scenarios.

\begin{table*}[t]
\centering
\caption{\textbf{Quantitative results on the DHP19 dataset.}}
\label{tab:result}
\begin{tabular}{l l c c c c}
\specialrule{1.2pt}{0pt}{0pt} 
Model & Method & MPJPE\textsubscript{2D}↓ & MPJPE\textsubscript{3D}↓ & PCK@0.4↑ & PCK@0.8↑ \\
\midrule
\multirow{3}{*}{CNN}
  & DHP19 \cite{[3HPE1]}  & 7.67 & 87.90 & - & -\\
  & LiftMono \cite{[3HPE4]}  & 26.79 & - & 0.28 & 0.51\\
  & MoveEnet $\dagger$ \cite{[2HPE1]}  & 6.52 & - & 0.97 & 0.98\\
\midrule
\multirow{2}{*}{PointNet}
  & Baseline $\dagger$ \cite{[3HPE5]}  & 7.34 & 83.12 & 0.859 & 0.976\\
  & Ours & \textbf{7.09} & \textbf{80.16} & \textbf{0.868} & \textbf{0.978}\\
\midrule
\multirow{2}{*}{DGCNN}
  & Baseline $\dagger$ \cite{[3HPE5]} & 6.85 & 77.68 & 0.88 & 0.982\\
  & Ours & \textbf{6.49} & \textbf{72.91} & \textbf{0.893} & \textbf{0.985}\\
\midrule
\multirow{2}{*}{Point Transformer}
  & Baseline $\dagger$ \cite{[3HPE5]} & 6.57 & 74.30 & 0.892 & 0.986\\
  & Ours & \textbf{6.38} & \textbf{72.16} & \textbf{0.898} & \textbf{0.988}\\
\specialrule{1.2pt}{0pt}{0pt} 
\end{tabular}
\end{table*}

\begin{table}[t]
\centering
\caption{\textbf{Complexity comparison between different input types.}}
\label{tab:complexity}
\begin{tabular}{l c c}
\specialrule{1.2pt}{0pt}{0pt}
Method & \#Params (M) & \#MACs (G) \\
\midrule
Pose-ResNet18 \cite{[simple]} & 15.4 & 8.30 \\
Pose-ResNet50 \cite{[simple]} & 34.0 & 12.91 \\
\midrule
PointNet (Ours) & 8.65 & 1.18 \\
DGCNN (Ours) & 8.70 & 4.81 \\
Point Transformer (Ours) & 4.70 & 5.04 \\
\specialrule{1.2pt}{0pt}{0pt}
\end{tabular}
\end{table}

\subsection{Implementation Details}
We follow the existing work \cite{[3HPE1]} by using subjects S1-S12 (training) and S13-S17 (testing). We adopt \cite{[3HPE5]} as the main baseline, as it is the most relevant work using sparse event point clouds, while other methods differ in data representation and are not directly comparable. The number of temporal slices is set to $K=4$, consistent with the rasterization process. All models are trained for 30 epochs using the Adam optimizer with a variable learning rate ($1\times10^{-4}$ initially, $1\times10^{-5}$ after 15 epochs, and $1\times10^{-6}$ after 20 epochs), a batch size of 32, and 2048 sampled points, on an RTX 5090 GPU.

We evaluate performance using Mean Per Joint Position Error (MPJPE), the average Euclidean distance between predicted and ground-truth joint positions in 2D (pixels) or 3D (millimeters), defined as:
\begin{equation}
\text{MPJPE} = \frac{1}{N \times J} \sum_{n=1}^{N} \sum_{j=1}^{J} \left\| \hat{\mathbf{p}}_{n,j} - \mathbf{p}_{n,j} \right\|_2,
\end{equation}
where $N$ is the number of samples, $J$ the number of keypoints, $\hat{\mathbf{p}}_{n,j}$ the prediction, and $\mathbf{p}_{n,j}$ the ground truth.

In addition to MPJPE, we employ the Percentage of Correct Keypoints (PCK) to further evaluate the robustness of our model. PCK measures the percentage of predicted keypoints that fall within a given threshold distance from the ground truth, defined as:
\begin{equation}
\text{PCK} = \frac{1}{N \times J} \sum_{n=1}^{N} \sum_{j=1}^{J} \mathbb{I}\left( \left| \hat{\mathbf{p}}{n,j} - \mathbf{p}{n,j} \right|_2 \le \tau \cdot L \right),
\end{equation}
where $\mathbb{I}(\cdot)$ is the indicator function, $\tau$ denotes the threshold, and $L$ is a normalization factor. In our experiments, we set the PCK thresholds to 0.4 and 0.8 to assess the localization accuracy under different levels of tolerance.

% \begin{table*}[t]
% \centering
% \caption{\textbf{Quantitative results on the DHP19 dataset.}}
% \label{tab:result}
% \begin{tabular}{l l c c c c}
% \hline
% Model & Method & MPJPE$_{2D}$$\downarrow$ & MPJPE$_{3D}$$\downarrow$ & PCK@0.4$\uparrow$ & PCK@0.8$\uparrow$ \\
% \hline
% \multirow{3}{*}{CNN}
%   & DHP19 \cite{[3HPE1]}  & 7.67 & 87.90 & - & -\\
%   & LiftMono \cite{[3HPE4]}  & 26.79 & - & 0.28 & 0.51\\
%   & MoveEnet $\dagger$ \cite{[2HPE1]}  & 6.52 & - & 0.97 & 0.98\\
  
% \hline
% \multirow{2}{*}{PointNet}
%   & Baseline $\dagger$ \cite{[3HPE5]}  & 7.34 & 83.12 & 0.859 & 0.976\\
%   & Ours & \textbf{7.09} & \textbf{80.16} & \textbf{0.868} & \textbf{0.978}\\
% \hline
% \multirow{2}{*}{DGCNN}
%   & Baseline $\dagger$ \cite{[3HPE5]} & 6.85 & 77.68 & 0.88 & 0.982\\
%   & Ours & \textbf{6.49} & \textbf{72.91} & \textbf{0.893} & \textbf{0.985}\\
% \hline
% \multirow{2}{*}{Point Transformer}
%   & Baseline $\dagger$ \cite{[3HPE5]} & 6.57 & 74.30 & 0.892 & 0.986\\
%   & Ours & \textbf{6.38} & \textbf{72.16} & \textbf{0.898} & \textbf{0.988}\\
% \hline
% \end{tabular}
% \end{table*}

\subsection{Comparative Results}

Table~\ref{tab:result} shows our results, with $\dagger$ indicating our reproduction, which shows similar results to the baseline. We also compare our point cloud-based framework with frame-based method (CNN) for event-driven HPE. As Table~\ref{tab:result} shows, our methods outperform the frame-based method in 2D MPJPE and 3D MPJPE. This indicates the potential of sparse point cloud representations to better preserve fine spatial details compared to dense frame-based methods, while maintaining computational efficiency.
For 3D Event Point Cloud input, the baseline uses three different point cloud backbones (PointNet, DGCNN, and Point Transformer).  We integrate our method into all three backbones, consistently improving their performance: PointNet reduces 2D and 3D MPJPE by about 3.4\% and 3.6\%, DGCNN achieves the largest improvement (5.3\% and 6.1\%), and Point Transformer improves by roughly 2.9\% in both metrics.
Notably, our DGCNN surpasses the baseline Point Transformer while having a simpler architecture and lower computational cost.
Furthermore, the results across PCK thresholds highlight our model's robustness. Under the PCK@0.4 and PCK@0.8 metric, our method consistently surpasses the point cloud baseline, with the Point Transformer backbone achieving a peak of 0.898 and 0.988.

We also evaluated the real-time performance on samples of 7,500 events (about 0.13 s each) with a batch size of 1.
PointNet and DGCNN achieve latencies of 1.89 ms and 3.73 ms, satisfying the requirement for real-time inference.
Table~\ref{tab:complexity} indicates the comparison of computational cost with frame-based methods, we follow the method \cite{[simple]}, where the Po
se-ResNet18 and Pose-ResNet50 models take a constant number (7500) of event frames as input.
In contrast, our point cloud-based models use fewer parameters and MACs, effectively reducing computational complexity.

\begin{table}[t]
\centering
\caption{\textbf{Ablation study on the proposed methods.}}
\label{tab:abaltion}
\begin{tabular}{l c c c}
\specialrule{1.2pt}{0pt}{0pt}
Method & MPJPE\textsubscript{2D}↓ & MPJPE\textsubscript{3D}↓ & PCK@0.4↑ \\
\midrule
Baseline & 7.34 & 83.12 & 0.859 \\
Baseline + S & 7.3 & 82.74 & 0.861 \\
Baseline + T & 7.12 & 80.56 & 0.864 \\
Baseline + T + S & 7.09 & 80.16 & 0.868 \\
\specialrule{1.2pt}{0pt}{0pt}
\end{tabular}
\end{table}

Fig.~\ref{fig:result} presents the visualization results. The figure shows challenges such as leg motion blur and unclear edges in static regions where no events are triggered. Our method produces predictions that better match the ground-truth skeleton and achieves lower MPJPE than the baseline. By explicitly modeling spatiotemporal properties through cross-slice dependencies, our method effectively resolves pose ambiguity in static or low-event scenarios, achieving a superior balance between accuracy and latency.

\begin{figure}[t]
  \centering
  \setlength{\tabcolsep}{1pt} 
  \renewcommand{\arraystretch}{0.9}

  \begin{tabular}{c c c c}
    % Baseline row
    \rotatebox{90}{\parbox{2cm}{\centering Baseline}} &
    \includegraphics[width=0.29\columnwidth]{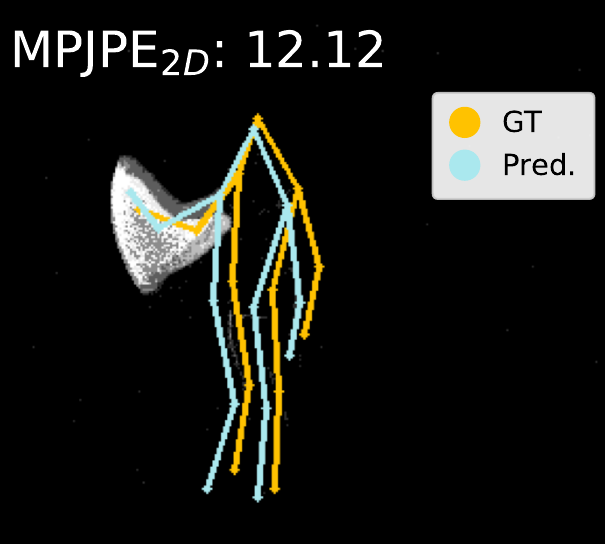} &
    \includegraphics[width=0.29\columnwidth]{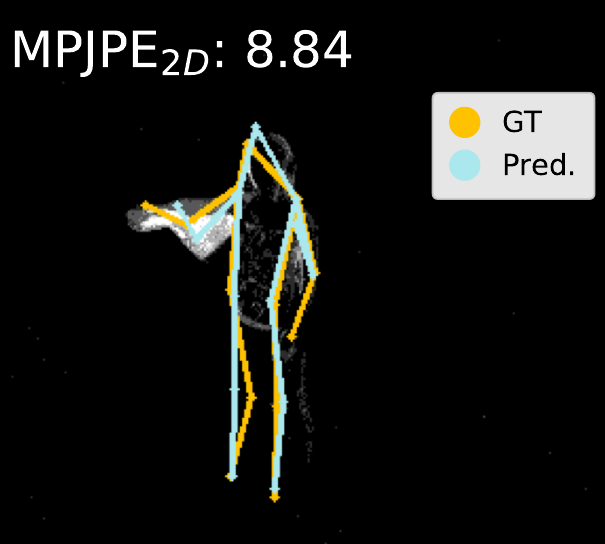} &
    \includegraphics[width=0.29\columnwidth]{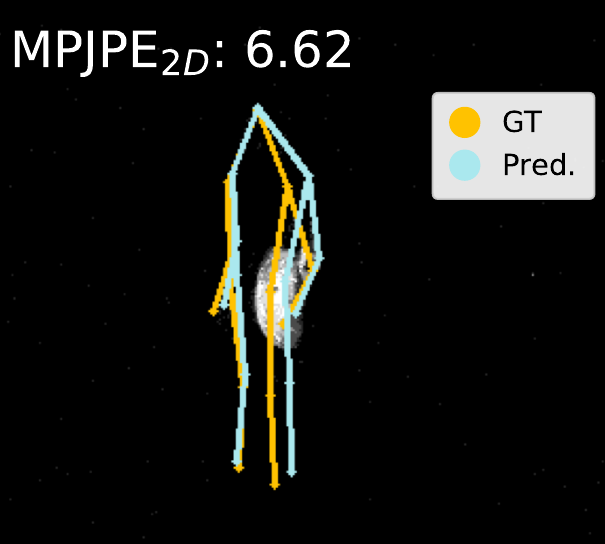} \\[-2pt]

    % Ours row
    \rotatebox{90}{\parbox{2cm}{\centering Ours}} &
    \includegraphics[width=0.29\columnwidth]{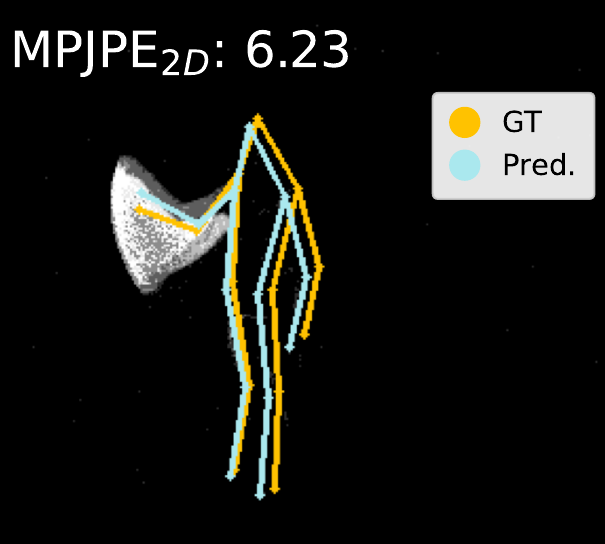} &
    \includegraphics[width=0.29\columnwidth]{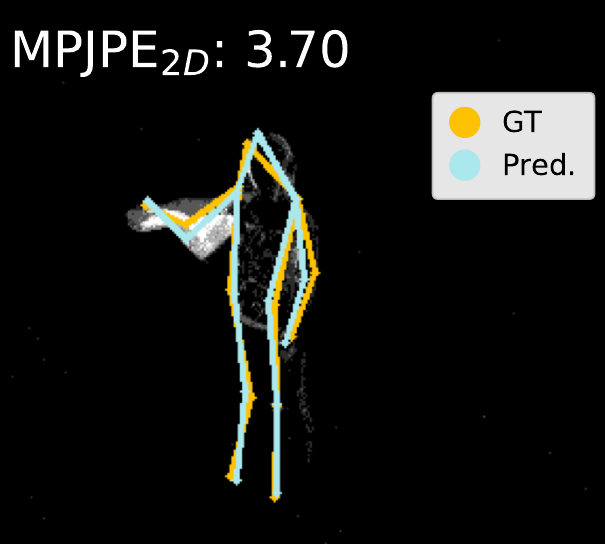} &
    \includegraphics[width=0.29\columnwidth]{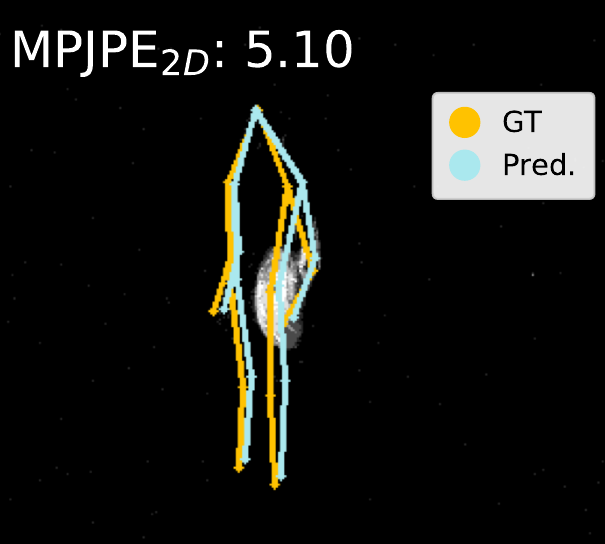} \\[-2pt]

    % (a)(b)(c)
    & \small (a) & \small (b) & \small (c)
  \end{tabular}

  \caption{Qualitative comparisons on the DHP19 dataset between the baseline and our method using the DGCNN backbone.}
  \label{fig:DGCNN}
\end{figure}

To further illustrate the advantages of the proposed method, Fig.~\ref{fig:DGCNN} provides qualitative comparisons using the DGCNN backbone. In the cases shown in (a) and (b), the subject’s left hand moves rapidly, producing sparse and discontinuous event patterns. The baseline model struggles to accurately localize the hand joint, while our method is able to track the motion more reliably and generate predictions that better match the ground truth. In (c), the subject remains relatively static, resulting in fewer triggered events. Even under such low-event conditions, our method still produces more accurate pose predictions, demonstrating its robustness in scenarios with limited event information.

\begin{figure}[t]
  \centering
  \setlength{\tabcolsep}{1pt} 
  \renewcommand{\arraystretch}{0.9}

  \begin{tabular}{c c c c}
    % Baseline row
    \rotatebox{90}{\parbox{2cm}{\centering Baseline}} &
    \includegraphics[width=0.29\columnwidth]{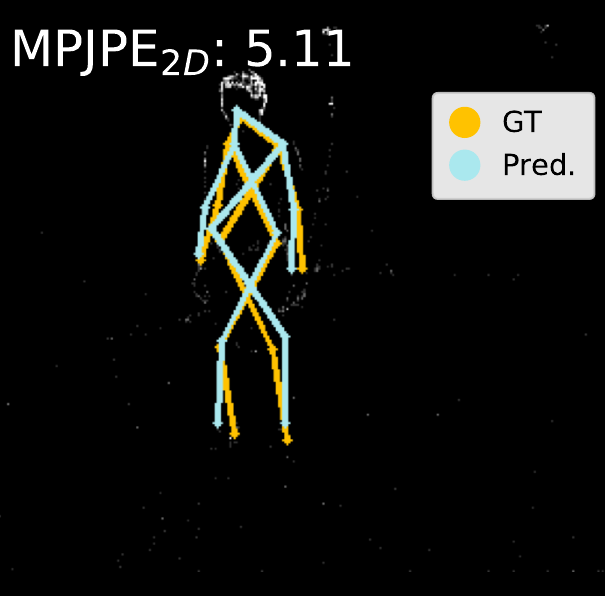} &
    \includegraphics[width=0.29\columnwidth]{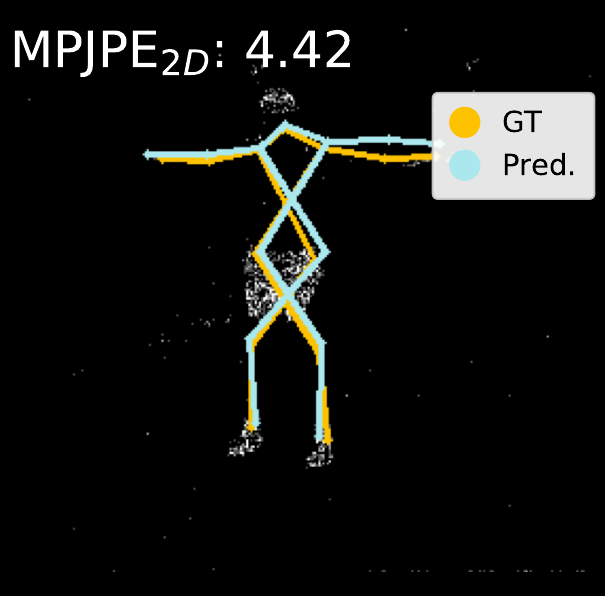} &
    \includegraphics[width=0.29\columnwidth]{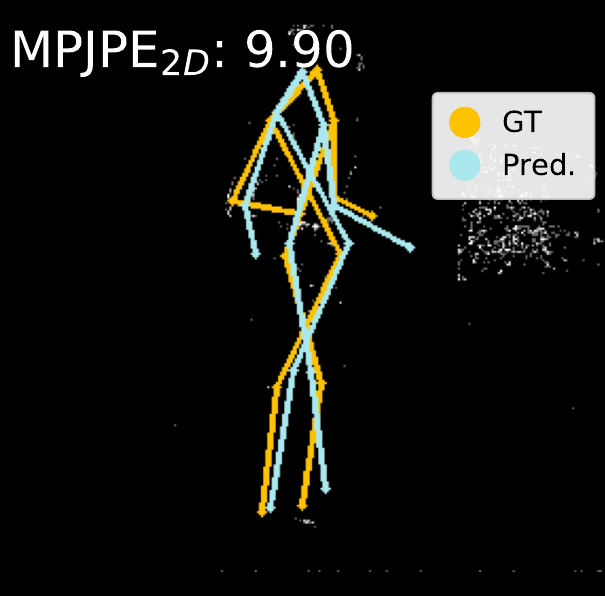} \\[-2pt]

    % Ours row
    \rotatebox{90}{\parbox{2cm}{\centering Ours}} &
    \includegraphics[width=0.29\columnwidth]{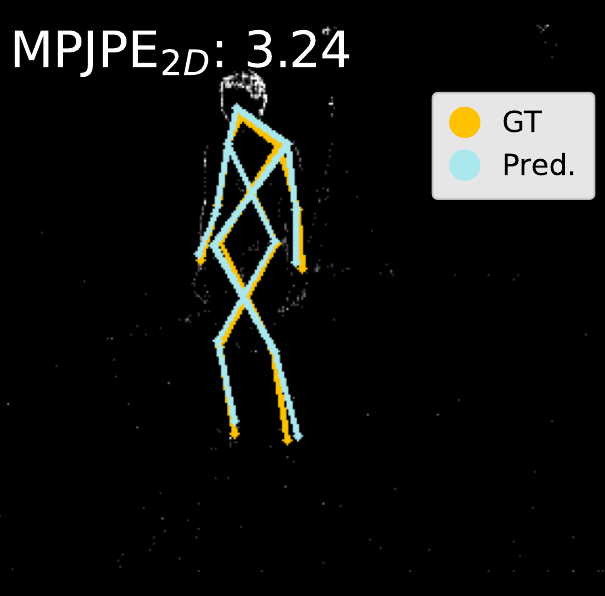} &
    \includegraphics[width=0.29\columnwidth]{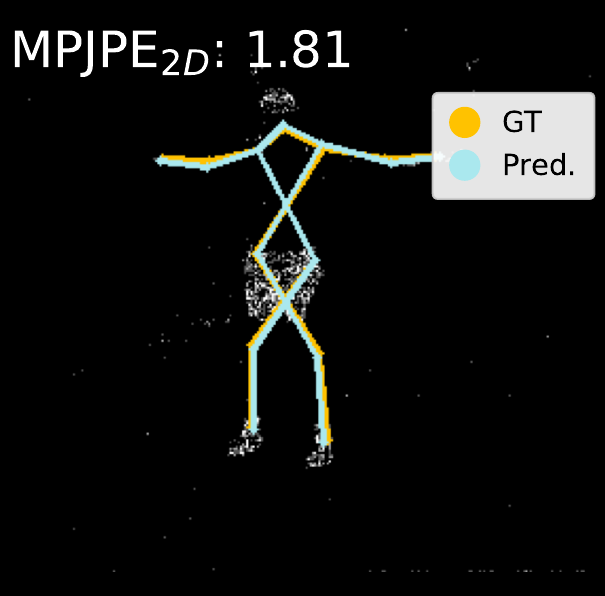} &
    \includegraphics[width=0.29\columnwidth]{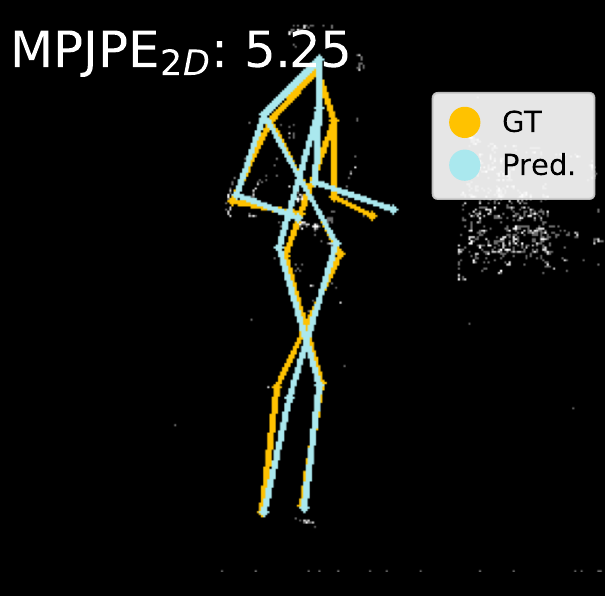} \\[-2pt]

    % (a)(b)(c)
    & \small (a) & \small (b) & \small (c)
  \end{tabular}

  \caption{Qualitative comparisons on the Event-Human3.6M dataset between the baseline and our method using the PointNet backbone.}
  \label{fig:h36m}
\end{figure}

We also present qualitative comparisons on the Event-Human3.6M dataset. As shown in Fig.~\ref{fig:h36m}, the predicted skeletons produced by our method align more closely with the ground-truth poses than the baseline across different motion scenarios. This result indicates that the proposed module can maintain reliable pose estimation performance under diverse event patterns, further demonstrating the effectiveness and robustness of our approach across different datasets.

% \begin{figure}[t]
%   \centering
%   \setlength{\tabcolsep}{1pt} 
%   \renewcommand{\arraystretch}{0.9}

%   \begin{tabular}{c c c}

%     \includegraphics[width=0.31\columnwidth]{h36m-figure/1.pdf} &
%     \includegraphics[width=0.31\columnwidth]{h36m-figure/2.pdf} &
%     \includegraphics[width=0.31\columnwidth]{h36m-figure/3.pdf} \\[-2pt]

%     \small (a) & \small (b) & \small (c)

%   \end{tabular}

%   \caption{Qualitative results on Event-Human3.6M using our method with the PointNet backbone.}
%   \label{fig:h36m}
% \end{figure}

\subsection{Ablation Study}

We performed an ablation study on PointNet to test our modules. As shown in Table~\ref{tab:abaltion}, when only the spatial Sobel-based edge enhancement (S) is introduced, the performance also improves compared to the baseline. Adding the temporal modeling module (T) improves both 2D and 3D pose estimation, reducing MPJPE$_{2D}$ and MPJPE$_{3D}$ by 3.0\% and 3.1\%. This shows that modeling temporal dependencies helps capture richer spatiotemporal features and improves keypoint accuracy. Combining both modules (T + S) achieves the best performance, further improving the results by about 0.4\%. This indicates that edge cues provide clearer structure to complement temporal modeling and enhance keypoint localization accuracy.

\section{Conclusion}

In this paper, we propose an efficient event point cloud-based human pose estimation framework that integrates temporal modeling (ES-Seq and ETSC) with spatial Sobel-based edge enhancement. Our method achieves consistent improvement across multiple backbones, with the improved DGCNN even surpassing the baseline Point Transformer while maintaining lower computational complexity. These results indicate that lightweight spatiotemporal modeling within a sparse point cloud pipeline can achieve a favorable balance between accuracy and efficiency. In future work, we plan to explore more adaptive and efficient spatiotemporal modeling strategies and extend the framework to broader event-based vision tasks.

%%%%%%%%%%%%%%%%%%%%%%%%%%%%%%%%%%%%%%%%%%%%%%%%%%%%%%%%%%%%%%%%%%%%%%%%%%%%%%%%

%%%%%%%%%%%%%%%%%%%%%%%%%%%%%%%%%%%%%%%%%%%%%%%%%%%%%%%%%%%%%%%%%%%%%%%%%%%%%%%%

%%%%%%%%%%%%%%%%%%%%%%%%%%%%%%%%%%%%%%%%%%%%%%%%%%%%%%%%%%%%%%%%%%%%%%%%%%%%%%%%

%%%%%%%%%%%%%%%%%%%%%%%%%%%%%%%%%%%%%%%%%%%%%%%%%%%%%%%%%%%%%%%%%%%%%%%%%%%%%%%%

\bibliographystyle{IEEEtran}
\bibliography{IEEEabrv, refs}

@article{[1],
  title={A survey of advances in vision-based human motion capture and analysis},
  author={Moeslund, Thomas B and Hilton, Adrian and Kr{\"u}ger, Volker},
  journal={Computer vision and image understanding},
  volume={104},
  number={2-3},
  pages={90--126},
  year={2006},
  publisher={Elsevier}
}

@inproceedings{[2],
  title={Simple baselines for human pose estimation and tracking},
  author={Xiao, Bin and Wu, Haiping and Wei, Yichen},
  booktitle={Proceedings of the European conference on computer vision (ECCV)},
  pages={466--481},
  year={2018}
}

@article{[3],
  title={Openpose: Realtime multi-person 2d pose estimation using part affinity fields},
  author={Cao, Zhe and Hidalgo, Gines and Simon, Tomas and Wei, Shih-En and Sheikh, Yaser},
  journal={IEEE transactions on pattern analysis and machine intelligence},
  volume={43},
  number={1},
  pages={172--186},
  year={2019},
  publisher={IEEE}
}

@article{[4],
  title={Event-based vision: A survey},
  author={Gallego, Guillermo and Delbr{\"u}ck, Tobi and Orchard, Garrick and Bartolozzi, Chiara and Taba, Brian and Censi, Andrea and Leutenegger, Stefan and Davison, Andrew J and Conradt, J{\"o}rg and Daniilidis, Kostas and others},
  journal={IEEE transactions on pattern analysis and machine intelligence},
  volume={44},
  number={1},
  pages={154--180},
  year={2020},
  publisher={IEEE}
}

@inproceedings{[5],
  title={Event-based vision meets deep learning on steering prediction for self-driving cars},
  author={Maqueda, Ana I and Loquercio, Antonio and Gallego, Guillermo and Garc{\'\i}a, Narciso and Scaramuzza, Davide},
  booktitle={Proceedings of the IEEE conference on computer vision and pattern recognition},
  pages={5419--5427},
  year={2018}
}

@article{[6],
  title={EV-FlowNet: Self-supervised optical flow estimation for event-based cameras},
  author={Zhu, Alex Zihao and Yuan, Liangzhe and Chaney, Kenneth and Daniilidis, Kostas},
  journal={arXiv preprint arXiv:1802.06898},
  year={2018}
}

@inproceedings{[7],
  title={End-to-end learning of representations for asynchronous event-based data},
  author={Gehrig, Daniel and Loquercio, Antonio and Derpanis, Konstantinos G and Scaramuzza, Davide},
  booktitle={Proceedings of the IEEE/CVF international conference on computer vision},
  pages={5633--5643},
  year={2019}
}

@inproceedings{[8],
  title={How late is too late? a preliminary event-based latency evaluation},
  author={Gava, Luna and Monforte, Marco and Bartolozzi, Chiara and Glover, Arren},
  booktitle={2022 8th International Conference on Event-Based Control, Communication, and Signal Processing (EBCCSP)},
  pages={1--4},
  year={2022},
  organization={IEEE}
}

@inproceedings{[3HPE1],
  title={DHP19: Dynamic vision sensor 3D human pose dataset},
  author={Calabrese, Enrico and Taverni, Gemma and Awai Easthope, Christopher and Skriabine, Sophie and Corradi, Federico and Longinotti, Luca and Eng, Kynan and Delbruck, Tobi},
  booktitle={Proceedings of the IEEE/CVF conference on computer vision and pattern recognition workshops},
  pages={0--0},
  year={2019}
}

@inproceedings{[3HPE2],
  title={Eventcap: Monocular 3d capture of high-speed human motions using an event camera},
  author={Xu, Lan and Xu, Weipeng and Golyanik, Vladislav and Habermann, Marc and Fang, Lu and Theobalt, Christian},
  booktitle={Proceedings of the IEEE/CVF Conference on Computer Vision and Pattern Recognition},
  pages={4968--4978},
  year={2020}
}

@inproceedings{[3HPE3],
  title={Eventhpe: Event-based 3d human pose and shape estimation},
  author={Zou, Shihao and Guo, Chuan and Zuo, Xinxin and Wang, Sen and Wang, Pengyu and Hu, Xiaoqin and Chen, Shoushun and Gong, Minglun and Cheng, Li},
  booktitle={Proceedings of the IEEE/CVF International Conference on Computer Vision},
  pages={10996--11005},
  year={2021}
}

@inproceedings{[3HPE4],
  title={Lifting monocular events to 3d human poses},
  author={Scarpellini, Gianluca and Morerio, Pietro and Del Bue, Alessio},
  booktitle={Proceedings of the IEEE/CVF Conference on Computer Vision and Pattern Recognition},
  pages={1358--1368},
  year={2021}
}

@inproceedings{[3HPE5],
  title={Efficient human pose estimation via 3d event point cloud},
  author={Chen, Jiaan and Shi, Hao and Ye, Yaozu and Yang, Kailun and Sun, Lei and Wang, Kaiwei},
  booktitle={2022 International Conference on 3D Vision (3DV)},
  pages={1--10},
  year={2022},
  organization={IEEE}
}

@inproceedings{[2HPE1],
  title={Moveenet: Online high-frequency human pose estimation with an event camera},
  author={Goyal, Gaurvi and Di Pietro, Franco and Carissimi, Nicolo and Glover, Arren and Bartolozzi, Chiara},
  booktitle={Proceedings of the IEEE/CVF Conference on Computer Vision and Pattern Recognition},
  pages={4024--4033},
  year={2023}
}

@article{[2HPE2],
  title={A temporal densely connected recurrent network for event-based human pose estimation},
  author={Shao, Zhanpeng and Wang, Xueping and Zhou, Wen and Wang, Wuzhen and Yang, Jianyu and Li, Youfu},
  journal={Pattern Recognition},
  volume={147},
  pages={110048},
  year={2024},
  publisher={Elsevier}
}

@article{[TCN],
  title={An empirical evaluation of generic convolutional and recurrent networks for sequence modeling},
  author={Bai, Shaojie and Kolter, J Zico and Koltun, Vladlen},
  journal={arXiv preprint arXiv:1803.01271},
  year={2018}
}

@article{[LSTM],
  title={Long short-term memory},
  author={Graves, Alex},
  journal={Supervised sequence labelling with recurrent neural networks},
  pages={37--45},
  year={2012},
  publisher={Springer}
}

@article{[o1],
  title={A Survey of 3D Reconstruction with Event Cameras},
  author={Xu, Chuanzhi and Zhou, Haoxian and Chen, Langyi and Chen, Haodong and Zhou, Ying and Chung, Vera and Qu, Qiang and Cai, Weidong},
  journal={arXiv preprint arXiv:2505.08438},
  year={2025}
}

@article{[o2],
  title={Ultralight Polarity-Split Neuromorphic SNN for Event-Stream Super-Resolution},
  author={Xu, Chuanzhi and Zhou, Haoxian and Chen, Langyi and Chung, Yuk Ying and Qu, Qiang},
  journal={arXiv preprint arXiv:2508.03244},
  year={2025}
}

@article{[o3],
  title={Towards End-to-End Neuromorphic Voxel-based 3D Object Reconstruction Without Physical Priors},
  author={Xu, Chuanzhi and Chen, Langyi and Chen, Haodong and Chung, Vera and Qu, Qiang},
  journal={arXiv preprint arXiv:2501.00741},
  year={2025}
}

@inproceedings{[simple],
  title={Simple baselines for human pose estimation and tracking},
  author={Xiao, Bin and Wu, Haiping and Wei, Yichen},
  booktitle={Proceedings of the European conference on computer vision (ECCV)},
  pages={466--481},
  year={2018}
}

@article{[SimDR],
  title={Is 2d heatmap representation even necessary for human pose estimation?},
  author={Li, Yanjie and Yang, Sen and Zhang, Shoukui and Wang, Zhicheng and Yang, Wankou and Xia, Shu-Tao and Zhou, Erjin},
  journal={CoRR},
  year={2021}
}

@article{[3HPE10],
  title={Highly Efficient 3D Human Pose Tracking from Events with Spiking Spatiotemporal Transformer},
  author={Zou, Shihao and Mu, Yuxuan and Ji, Wei and Wang, Zi-An and Zuo, Xinxin and Wang, Sen and Si, Weixin and Cheng, Li},
  journal={IEEE Transactions on Circuits and Systems for Video Technology},
  year={2025},
  publisher={IEEE}
}

@article{[SNN],
  title={Spiking neural networks},
  author={Ghosh-Dastidar, Samanwoy and Adeli, Hojjat},
  journal={International journal of neural systems},
  volume={19},
  number={04},
  pages={295--308},
  year={2009},
  publisher={World Scientific}
}

@inproceedings{[HPE1],
  title={Deeppose: Human pose estimation via deep neural networks},
  author={Toshev, Alexander and Szegedy, Christian},
  booktitle={Proceedings of the IEEE conference on computer vision and pattern recognition},
  pages={1653--1660},
  year={2014}
}

@inproceedings{[HPE2],
  title={3d human pose estimation in video with temporal convolutions and semi-supervised training},
  author={Pavllo, Dario and Feichtenhofer, Christoph and Grangier, David and Auli, Michael},
  booktitle={Proceedings of the IEEE/CVF conference on computer vision and pattern recognition},
  pages={7753--7762},
  year={2019}
}

@inproceedings{[HPE3],
  title={Tokenpose: Learning keypoint tokens for human pose estimation},
  author={Li, Yanjie and Zhang, Shoukui and Wang, Zhicheng and Yang, Sen and Yang, Wankou and Xia, Shu-Tao and Zhou, Erjin},
  booktitle={Proceedings of the IEEE/CVF International conference on computer vision},
  pages={11313--11322},
  year={2021}
}

@inproceedings{[HPE4],
  title={Diffpose: Multi-hypothesis human pose estimation using diffusion models},
  author={Holmquist, Karl and Wandt, Bastian},
  booktitle={Proceedings of the IEEE/CVF international conference on computer vision},
  pages={15977--15987},
  year={2023}
}

@inproceedings{[HPE5],
  title={Unipose: A unified multimodal framework for human pose comprehension, generation and editing},
  author={Li, Yiheng and Hou, Ruibing and Chang, Hong and Shan, Shiguang and Chen, Xilin},
  booktitle={Proceedings of the Computer Vision and Pattern Recognition Conference},
  pages={27805--27815},
  year={2025}
}

@inproceedings{[action],
  title={3DWSNet: A Novel 3D Wavelet Spiking Neural Network for Event-based Action Recognition},
  author={Fang, Junkang and Dang, Yonghao and Zhao, Wending and Yu, Bo and Wang, Zehao and Yin, Jianqin},
  booktitle={2025 IEEE/RSJ International Conference on Intelligent Robots and Systems (IROS)},
  pages={11607--11613},
  year={2025},
  organization={IEEE}
}

@inproceedings{[tracking],
  title={6-DoF Object Tracking with Event-based Optical Flow and Frames},
  author={Li, Zhichao and Glover, Arren and Bartolozzi, Chiara and Natale, Lorenzo},
  booktitle={2025 IEEE/RSJ International Conference on Intelligent Robots and Systems (IROS)},
  pages={18880--18887},
  year={2025},
  organization={IEEE}
}

@inproceedings{[HPE6],
  title={Stacked hourglass networks for human pose estimation},
  author={Newell, Alejandro and Yang, Kaiyu and Deng, Jia},
  booktitle={European conference on computer vision},
  pages={483--499},
  year={2016},
  organization={Springer}
}

@inproceedings{[HPE7],
  title={Realtime multi-person 2d pose estimation using part affinity fields},
  author={Cao, Zhe and Simon, Tomas and Wei, Shih-En and Sheikh, Yaser},
  booktitle={Proceedings of the IEEE conference on computer vision and pattern recognition},
  pages={7291--7299},
  year={2017}
}

@article{[h36m],
  title={Human3. 6m: Large scale datasets and predictive methods for 3d human sensing in natural environments},
  author={Ionescu, Catalin and Papava, Dragos and Olaru, Vlad and Sminchisescu, Cristian},
  journal={IEEE transactions on pattern analysis and machine intelligence},
  volume={36},
  number={7},
  pages={1325--1339},
  year={2013},
  publisher={IEEE}
}

@inproceedings{[v2e],
  title={v2e: From video frames to realistic DVS events},
  author={Hu, Yuhuang and Liu, Shih-Chii and Delbruck, Tobi},
  booktitle={Proceedings of the IEEE/CVF conference on computer vision and pattern recognition},
  pages={1312--1321},
  year={2021}
}

\end{document}